\documentclass[10pt,twocolumn,letterpaper]{article}

\PassOptionsToPackage{table}{xcolor}
\usepackage[pagenumbers]{cvpr}
\usepackage[T1]{fontenc}
\usepackage{fix-cm}
\usepackage{multirow}
\usepackage{tabularx}
\usepackage{placeins}
\usepackage{microtype}
\definecolor{resultblue}{RGB}{236,243,248}
\definecolor{cvprblue}{rgb}{0.21,0.49,0.74}
\usepackage[pagebackref,breaklinks,colorlinks,allcolors=cvprblue]{hyperref}
\hypersetup{
  pdftitle={CoT-WAM: Collaborative Training of Vision-Language Models and World-Action Models via Plan Latents},
  pdfauthor={Jinhao Zhang, Yushen Zuo, Ke Rui, Weitao Zhou, Haibao Yu, Honglong Tian, Xiaoheng Ma, Jie Mei, Jiawei Wang, Minglei Li},
  pdfsubject={Robot learning; vision-language models; world-action models},
  pdfkeywords={CoT-WAM, robot manipulation, world-action models, visual planning}
}

\title{CoT-WAM: Collaborative Training of Vision-Language Models\\
and World-Action Models via Plan Latents}
\author{%
\parbox{0.95\textwidth}{\centering
\fontsize{11}{13}\selectfont
Jinhao Zhang$^{1}$\quad
Yushen Zuo$^{2}$\quad
Ke Rui$^{2}$\quad
Weitao Zhou$^{2,3,\dagger,\ddagger}$\quad
Haibao Yu$^{4,5,\ddagger}$\\[4pt]
Honglong Tian$^{2}$\quad
Xiaoheng Ma$^{6}$\quad
Jie Mei$^{1,\ddagger}$\quad
Jiawei Wang$^{2}$\quad
Minglei Li$^{2}$\\[8pt]
\small
$^{1}$Harbin Institute of Technology, Shenzhen\qquad $^{2}$Simple AI\\[2pt]
$^{3}$Tsinghua University\qquad $^{4}$Tuojing Intelligence\qquad $^{5}$The University of Hong Kong\\[2pt]
$^{6}$The Hong Kong University of Science and Technology\\[5pt]
$^{\dagger}$Project lead.\qquad $^{\ddagger}$Corresponding authors.
}}

\newcommand{\method}{CoT-WAM}

\makeatletter
\patchcmd{\@maketitle}{\vskip .375in}{\vskip .10in}{}{}
\patchcmd{\@maketitle}{\vspace*{24pt}}{\vspace*{10pt}}{}{}
\patchcmd{\@maketitle}{\vspace*{12pt}}{\vspace*{4pt}}{}{}
\makeatother

\begin{document}
\raggedbottom
\maketitle

\begin{abstract}
General robot manipulation requires both semantic competence and physical foresight. World action models (WAMs) acquire an understanding of and predictive capability over the physical world by jointly modeling visual foresight and action distributions. However, existing WAMs still lack semantic competence: they have limited ability to perform open-world scene understanding, semantic reasoning, and long-horizon planning. Although vision-language models (VLMs) provide these capabilities, effectively integrating their semantic priors into WAM prediction and control remains challenging. To address this challenge, we propose \textbf{CoT-WAM}, a jointly trained WAM--VLM framework. CoT-WAM uses a learnable embedding to exchange information between the VLM and the WAM, which is supervised by textual and goal-image signals during training to encode rich intent information. A tailored two-stage training strategy enables joint optimization and more effective collaboration between the VLM and WAM. We evaluate CoT-WAM on six benchmarks covering control performance, zero-shot generalization, and long-horizon planning. CoT-WAM achieves SOTA performance on LIBERO, CALVIN, and RoboCasa-Kitchen, and demonstrates stronger zero-shot generalization on LIBERO-Plus and LIBERO-Pro without fine-tuning. On RoboCasa365, it further improves long-horizon planning over baselines with and without external planners. Real-world experiments further demonstrate the effectiveness of our method in practical deployments. Code is available at: \url{https://github.com/jhz1192/CoT-WAM.git}
\end{abstract}

\vspace{-7pt}
\section{Introduction}
\label{sec:intro}

Vision-language-action models (VLAs) have recently emerged as a dominant paradigm in embodied intelligence~\citep{brohan2023rt2visionlanguageactionmodelstransfer,kim2024openvlaopensourcevisionlanguageactionmodel,black2026pi0visionlanguageactionflowmodel}. By transferring the powerful semantic and scene-understanding capabilities of vision-language models (VLMs) to robot control signals through discrete action tokens or dedicated action experts, VLAs achieve strong generalization. However, their reactive control paradigm cannot model the evolution of the physical world or predict the consequences of actions, both of which are essential capabilities for open-world manipulation agents. World Action Models (WAMs) address this limitation by jointly modeling future visual-state transitions and action distributions~\citep{li2025unifiedvideoactionmodel,yuan2026fastwamworldactionmodels,li2026causalworldmodelingrobot}. By conditioning action generation on learned world dynamics, WAMs provide a viable paradigm for pretraining on large-scale videos and make it possible to build manipulation agents capable of understanding the physical world.

Despite their ability to predict state evolution, WAMs lack the open-world understanding, reasoning, and planning capabilities provided by a VLM backbone. Existing WAMs, such as Fast-WAM~\citep{yuan2026fastwamworldactionmodels}, LingBot-VA~\citep{li2026causalworldmodelingrobot}, and DreamZero~\citep{ye2026worldactionmodelszeroshot}, typically rely on a separate frozen text encoder to provide language grounding. This design is often insufficient for tasks that require open-world scene understanding, semantic generalization and long-horizon planning. Recent works, such as ST-WAM~\citep{wang2026stwamsemantictemporalworldaction} and SG-WAM~\citep{he2026sgwamtextgroundedspatialawaresemantic}, incorporate VLMs to provide scene-aware semantic guidance for action generation. However, ST-WAM keeps its VLM frozen throughout training, so its semantic representations are not jointly optimized with the WAM. This limits the adaptation of the VLM's semantic knowledge to the needs of future-state prediction and action generation. SG-WAM jointly trains a VLM-based semantic planner with the WAM, but relies on additional frozen semantic and geometric teachers to supervise future feature prediction.

To address these limitations, we propose \textbf{CoT-WAM}, a framework that seamlessly integrates the rich semantic prior of a VLM into a WAM, thereby equipping WAM prediction and decision-making with open-world understanding and planning capabilities. Building upon Fast-WAM, CoT-WAM replaces its conditioning input with a learnable hidden state produced by the VLM. The VLM is responsible for scene understanding or subtask planning, and conveys its planning intent to the WAM through the latent embedding. Conditioned on this compact yet semantically rich representation, the WAM predicts the corresponding future states and generates control signals. During training, we supervise the learnable embedding with two complementary signals: a discrete textual task signal and a continuous goal-image signal. Consequently, the information transmitted from the VLM to the WAM is no longer restricted to ``hard'' text, but can encode richer information about the predicted consequences of an intended action. Through a two-stage training strategy, we successfully endow the WAM with stronger understanding and planning capabilities, leading to improved closed-loop control performance.

We evaluate CoT-WAM across six benchmarks covering control performance, zero-shot generalization, and long-horizon planning. For control, CoT-WAM achieves state-of-the-art performance on LIBERO, CALVIN, and RoboCasa-Kitchen without additional embodied pretraining. For zero-shot generalization, the same LIBERO-trained policy substantially outperforms Fast-WAM on LIBERO-Plus and LIBERO-Pro without fine-tuning, while remaining competitive with methods using additional embodied pretraining or supervision from external visual priors. For long-horizon planning, CoT-WAM outperforms baselines with and without external planners on RoboCasa365, a challenging benchmark for mobile manipulation. Beyond these simulation benchmarks, real-world experiments further validate the practicality of CoT-WAM. Together, these results demonstrate that integrating VLM semantic knowledge with world modeling through joint training improves scene understanding, semantic generalization, and long-horizon planning. In summary, our contributions are:
\begin{itemize}
    \item We propose \textbf{CoT-WAM}, a joint WAM--VLM training framework that equips WAMs with scene understanding, semantic generalization, and long-horizon planning capabilities.
    \item We introduce dedicated latent tokens to enable richer and more flexible information exchange between the VLM and WAM, supervised by text and goal images. A staged training strategy enables joint learning and collaboration between the two models.
    \item Experiments across six benchmarks demonstrate that CoT-WAM substantially improves long-horizon planning over the baseline, while achieving state-of-the-art control performance and stronger zero-shot generalization.
\end{itemize}

\section{Related Work}
\label{sec:related}

\subsection{Vision-Language-Action Models}
Vision-language-action models transfer semantic knowledge from pretrained VLMs to robot control. OpenVLA~\citep{kim2024openvlaopensourcevisionlanguageactionmodel} learns manipulation policies from multi-robot demonstrations, while $\pi_0$~\citep{black2026pi0visionlanguageactionflowmodel} couples a VLM with flow-matching action generation. X-VLA~\citep{zheng2025xvlasoftpromptedtransformerscalable} studies scalable adaptation across embodiments. $\pi_{0.5}$~\citep{intelligence2025pi05visionlanguageactionmodelopenworld} combines robot data, web data, and semantic subtask prediction to improve open-world generalization. These methods emphasize semantic understanding and action generation without explicitly supervising future visual-state prediction. We complement such semantic priors with WAM-based visual prediction, so that the representations used for control are also trained to capture task-relevant scene evolution.

\subsection{World Action Models}
World action models learn visual dynamics alongside robot actions. LingBot-VA~\citep{li2026causalworldmodelingrobot} couples video and action experts through causal modeling. Fast-WAM~\citep{yuan2026fastwamworldactionmodels} retains future-video prediction as training supervision while omitting it at inference, and Faster-WAM~\citep{ma2026fasterwamworldactionmodels} further reduces computation with a lightweight action module. However, language conditioning in models such as Fast-WAM is encoded independently of the current observation, leaving scene-dependent task interpretation to the downstream video--action network. Building on Fast-WAM, we replace this conditioning interface with continuous intent representations produced by a VLM from both the observation and instruction, providing scene-aware semantic guidance for visual prediction and action generation.

\subsection{Integrating VLMs with World Action Models}
Recent work incorporates VLM priors into WAMs through different conditioning interfaces. DSWAM~\citep{zhu2026dswamdualsystemworldaction} introduces an optional VLM planner that supplies explicit subtask instructions to a WAM executor, but this discrete text interface does not provide a differentiable path from the executor to the planner. ST-WAM~\citep{wang2026stwamsemantictemporalworldaction} uses frozen VLM features to retrieve task-relevant semantic history for action generation, so WAM training cannot adapt the VLM itself. SG-WAM~\citep{he2026sgwamtextgroundedspatialawaresemantic} enables joint training of a VLM-based planner and the WAM, but requires additional frozen semantic and geometric teachers to supervise future feature prediction. Our approach jointly optimizes the VLM and video predictor through a continuous interface supervised by task text and future visual observations, without additional external priors. Predicted visual goals further ground the communicated intent in a future state that conditions both video and action generation.

\section{Overview}

We consider embodied policy learning from visual observations and language instructions. Given the current observation $o_t$, robot proprioceptive state $s_t$, and task instruction $\ell$, the policy predicts an action sequence of length $H$, $A_t=(a_t,\ldots,a_{t+H-1})$. Following Fast-WAM~\citep{yuan2026fastwamworldactionmodels}, we use future video prediction as auxiliary supervision to shape visual representations during training, while omitting full video prediction at inference time to accelerate execution.

\begin{figure*}[t]
\centering
\includegraphics[page=1,width=\linewidth]{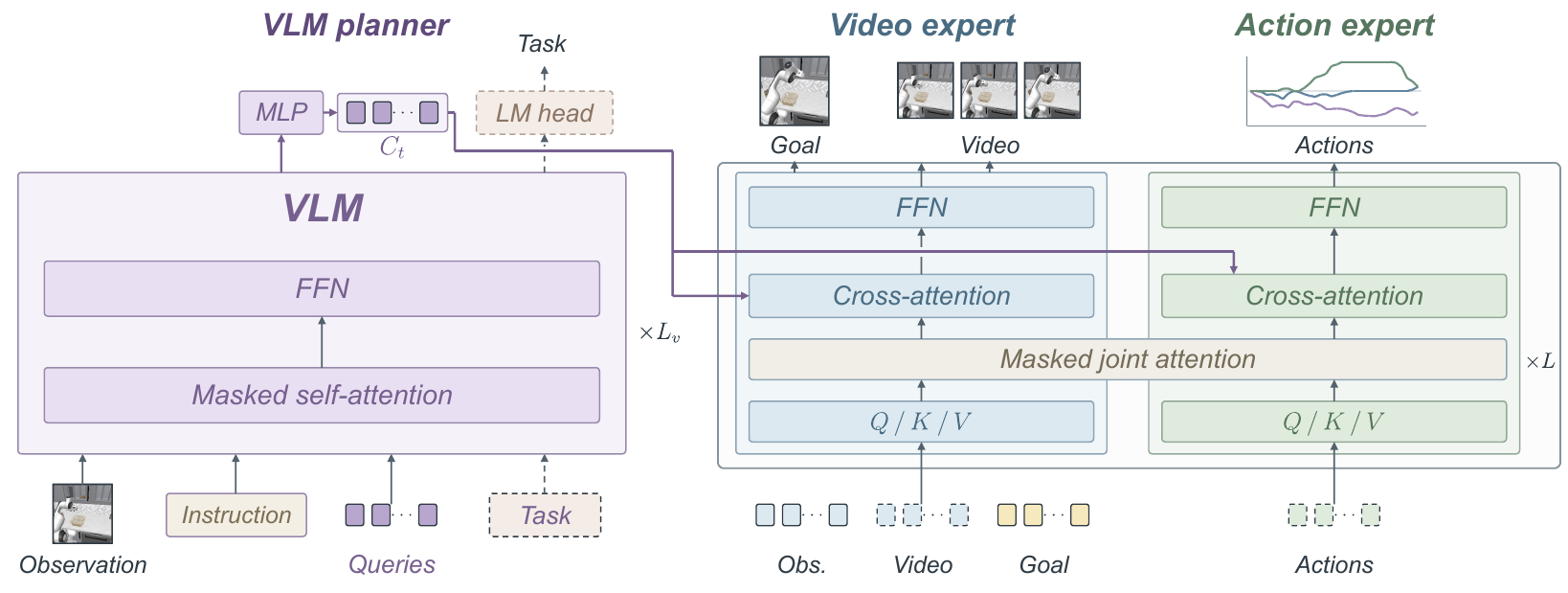}
\caption{\textbf{Overview of \method{}.} The VLM encodes the current observation and instruction into continuous intent tokens $C_t$. The world action model uses these tokens to predict a visual goal, which conditions the video and action experts. Text and visual supervision train the interface in Stage 1; Stage 2 learns control with the VLM frozen. Full future-video generation is omitted during action inference.}
\label{fig:architecture}
\end{figure*}

As shown in Figure~\ref{fig:architecture}, our model consists of a world action model based on a MoT architecture~\citep{liang2025mixtureoftransformerssparsescalablearchitecture} and a VLM planner. Given the current observation and task, the VLM planner communicates its plan for the next step to the world action model through a latent interface. The world action model then generates the action chunk $A_t$ conditioned on the VLM's intent and the current observation. To make the intent signal transmitted from the VLM to the world model more concrete and continuous, we further introduce goal-image prediction on the video-model side, mapping the VLM's continuous hidden representations to a concrete target visual state. The predicted goal image then conditions both action and video generation, grounding the VLM's intent more directly in control. This grounds the WAM's control outputs in the VLM's strong semantic priors, improving closed-loop performance and generalization.

\section{Method}
\label{sec:method}

\begin{figure}[!htbp]
\centering
\includegraphics[width=0.95\linewidth]{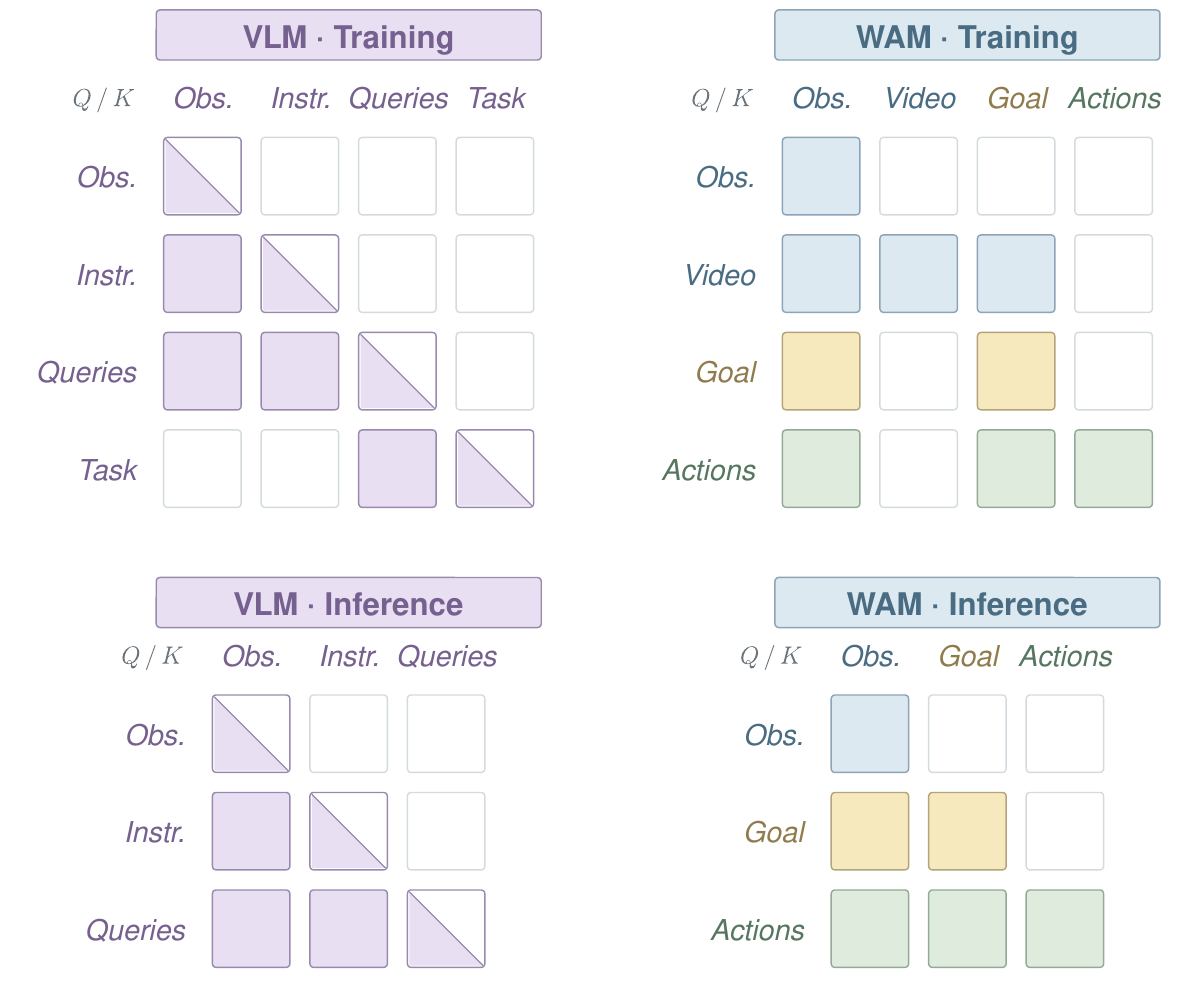}
\caption{\textbf{Training and inference attention masks.} Rows are queries and columns are keys. Colored cells allow attention; triangles denote causal attention. Inference omits the VLM task-text tokens and WAM future-video tokens, while retaining the latent interface and goal-conditioned action path.}
\label{fig:attention-mask}
\end{figure}

\subsection{Latent VLM--WAM Interface}
\label{sec:latent-interface}

To better incorporate the VLM's understanding and planning of the current scene and task into action generation, we introduce a learnable latent embedding interface for communication between the VLM and the WAM. Specifically, we encode the current observation $o_t$ and task instruction $\ell$, then append $K$ learnable intent tokens $Q=(q_1,\ldots,q_K)$ to form the VLM input sequence. Through the VLM's self-attention, these tokens aggregate scene and task information. Their final hidden states form a continuous representation of the VLM's intent, which is passed to the downstream WAM through a projection layer:
\begin{equation}
Z_t=F_{\phi}(o_t,\ell,Q)\big|_{Q},
\qquad C_t=P_{\psi}(Z_t),
\label{eq:interface}
\end{equation}
where $F_{\phi}$ denotes the VLM and $P_{\psi}$ is an MLP projection layer.

The downstream WAM adopts a MoT architecture similar to Fast-WAM~\citep{yuan2026fastwamworldactionmodels}, in which a video expert and an action expert are integrated through joint attention. Both experts access the intent information $C_t$ from the VLM through cross-attention to condition future-state and action generation. This communication mechanism allows future prediction gradients to propagate through the interface to the VLM, enabling joint training and closer collaboration between the VLM and WAM.

\subsection{Latent Interface Supervision}
\label{sec:latent-supervision}

The differentiable interface establishes an information pathway between the VLM and the WAM, but appropriate training objectives are still needed to determine what information it should convey. We therefore introduce two complementary forms of supervision: text and goal images. Text supervision encourages the interface to retain high-level task semantics or subtask planning objectives, while goal-image supervision explicitly expresses the VLM's intent in visual form, enriching the transmitted information and grounding it more directly in control.

\paragraph{Task Text Supervision}
Unlike previous methods that introduce an auxiliary language model for text supervision~\citep{li2026wallwmcarvingworldaction}, we use the VLM planner's own text generation capability. Specifically, during training, we append the target text $y_t$ after the intent tokens $Q$ and train the VLM planner to predict it autoregressively:
\begin{equation}
\mathcal{L}_{\mathrm{text}}
=-\mathbb{E}_{\mathcal D}\sum_{j=1}^{|y_t|}
\log p_{\phi}(y_{t,j}\mid y_{t,<j},o_t,\ell,Q).
\label{eq:text-loss}
\end{equation}
Here, the target text can be set to detailed subtask annotations or reasoning text, or it can simply be the full task prompt.

To ensure that the information used for text prediction comes entirely through the latent queries, we design an attention mask that restricts the target text to reading the latent queries while blocking its attention to the observation images and task instruction. This forces the latent queries to encode information relevant to subtask prediction or the semantics of the task. In our implementation, we use the task prompt directly as the target text for general manipulation tasks, while for long-horizon planning tasks, we use the provided subtask descriptions as prediction targets. At inference time, we obtain the VLM's intent output $C_t$ in a single forward pass without decoding the subsequent text output. The attention masks used during training and inference are shown in Figure~\ref{fig:attention-mask}.

\paragraph{Goal-Image Supervision}

Supervision based only on task semantics or subtask planning is often too coarse to make the latent queries encode the VLM's rich scene understanding and specific intent. We therefore additionally supervise the latent interface using the image at the end of the training window, $g_t=o_{t+H}$, as the goal image. As with text supervision, we use the capabilities of the existing module: the WAM's video expert provides goal-image supervision through its own visual prediction capability. Specifically, we train the video expert to predict the goal image from the current observation and the intent condition $C_t$ given by the VLM. This forces the latent interface to encode the concrete visual changes associated with the intent, grounding the VLM's abstract intent in a specific visual state.

\subsection{Goal-Conditioned World-Action Modeling}
\label{sec:world-action}

For the WAM module, we adopt the basic architecture of Fast-WAM~\citep{yuan2026fastwamworldactionmodels}. The video and action experts predict visual evolution and action sequences, respectively, and access the intent information $C_t$ from the VLM through cross-attention as a prediction condition. To better ground action and future-video prediction in the VLM's visual intent, we additionally use the predicted goal image to condition both action and video generation. During training, we predict the goal image $g_t$ from the current observation $o_t$ and the latent condition $C_t$ given by the VLM, and predict action sequences and future video conditioned on $o_t,g_t,C_t$. As in Fast-WAM, action generation is prevented from attending to future video. Future prediction therefore serves only as auxiliary supervision for representation learning during training, while inference skips explicit prediction of the full future video trajectory for efficiency. Figure~\ref{fig:attention-mask} illustrates the corresponding attention mask.

During training, a frozen VAE encodes goal images and future video into latent space, where they are modeled generatively alongside action sequences. All three prediction tasks use the same flow-matching formulation~\citep{lipman2023flowmatchinggenerativemodeling}. For a prediction target $y$, we sample Gaussian noise $\epsilon\sim\mathcal{N}(0,I)$ and construct a noisy input at noise timestep $\tau \in (0,1)$:

\begin{equation}
y_{\tau}=(1-\tau)y+\tau\epsilon.
\label{eq:flow-path}
\end{equation}
The model predicts a velocity from the noisy input and the corresponding generation condition $c$, with the training objective
\begin{equation}
\mathcal{L}_{\mathrm{FM}}(y)
=
\mathbb{E}_{y,\epsilon,\tau}
\left[
\left\|
f_{\theta}(y_{\tau},\tau;c)-(\epsilon-y)
\right\|_2^2
\right].
\label{eq:flow-loss}
\end{equation}
Instantiating $y$ as the goal image, future video, or action sequence gives the respective losses $\mathcal{L}_{\mathrm{goal}}$, $\mathcal{L}_{\mathrm{video}}$, and $\mathcal{L}_{\mathrm{action}}$. Goal-image and video generation share the video expert, while action generation uses the action expert. All three prediction tasks can access the current observation and the intent condition supplied by the VLM; video and action prediction additionally access the goal image as a condition. At inference time, we first predict the goal image from the latent queries and then use it to condition action prediction, making goal prediction an intermediate representation that connects semantic understanding with control.

\subsection{Staged Training and Closed-Loop Inference}
\label{sec:training-inference}
\subsubsection{Staged Training Strategy}
We divide semantic interface learning and action policy learning into two stages. The first stage jointly learns intent representations and visual prediction, enabling VLM outputs to serve as effective generation conditions for the WAM. The second stage fixes this interface and learns action generation while retaining visual supervision.

\paragraph{Supervising the Latent Interface.}
We first supervise how the latent interface encodes information. The latent queries should capture both (1) task or subtask semantics and (2) the target visual state and its evolution. We therefore jointly train the VLM and the video expert in this stage, without incorporating the action expert. Specifically, we optimize
\begin{equation}
\mathcal L_{\mathrm{stage_1}}
=\lambda_{\mathrm{text}}\mathcal L_{\mathrm{text}}
+\lambda_{\mathrm{goal}}\mathcal L_{\mathrm{goal}}
+\lambda_{\mathrm{video}}\mathcal L_{\mathrm{video}}.
\label{eq:stage1}
\end{equation}
Text supervision encourages the interface to preserve task semantics or subtask planning information, while the goal and video losses jointly propagate visual prediction gradients to the VLM, encouraging the latent queries to encode continuous visual-state evolution. This stage teaches the latent queries to encode the information needed for downstream prediction.

\paragraph{Joint Optimization of Action and Video Generation.} In Stage 2, we load all weights trained in Stage 1 and initialize the action expert by linearly interpolating the video expert's weights. We freeze the VLM and the conditioning projection layer, and jointly optimize the video and action experts:
\begin{equation}
\mathcal L_{\mathrm{stage_2}}
=\mathcal L_{\mathrm{goal}}
+\mathcal L_{\mathrm{video}}
+\mathcal L_{\mathrm{action}}.
\label{eq:stage2}
\end{equation}
Conditioned on the intent supplied by the frozen VLM, this stage jointly trains the entire downstream WAM module to translate the VLM's abstract planning intent into concrete control signals.

\subsubsection{Closed-Loop Inference}
At inference time, the VLM produces an intent condition $C_t$ from the latest observation and task instruction and passes it to the downstream WAM. The video expert first predicts a goal image $\hat g_t$ from the current observation $o_t$ and $C_t$. The action expert then additionally conditions on this goal image to generate the next action chunk $\hat A_t$:
\begin{equation}
\hat g_t=f_g(o_t,C_t),\qquad
\hat A_t=f_a(o_t,C_t,\hat g_t).
\label{eq:inference}
\end{equation}
Here, $f_g$ and $f_a$ denote the video and action experts, respectively. During training, we add noise to the goal condition with probability 0.5 so that the action expert can adapt to noisy goal representations:
\begin{equation}
\begin{gathered}
\tilde{g}_t =
\begin{cases}
g_t, & \text{w.p. } \tfrac{1}{2}, \\[2pt]
(1-\sigma)g_t+\sigma\epsilon, & \text{o.w. }
\end{cases}
\end{gathered}
\label{eq:goal-condition-noise}
\end{equation}
where $\sigma\sim\mathcal{U}(0,0.5)$ and $\epsilon\sim\mathcal{N}(0,I)$. At inference time, we therefore denoise the goal image only to half the timestep and then fix it as the condition for action generation, accelerating inference.

\section{Experiments}
\label{sec:experiments}

\begin{table*}[t]
\centering
\caption{\textbf{Control performance.} Success rate (\%) on LIBERO/RoboCasa-Kitchen; average successful chain length (ACL) on CALVIN. Emb. PT.: additional embodied pretraining. Bold/underline: group best/second. $\Delta$: gains over Fast-WAM baseline.}
\label{tab:main}
\small
\setlength{\tabcolsep}{4pt}
\setlength{\aboverulesep}{1pt}
\setlength{\belowrulesep}{1pt}
\renewcommand{\arraystretch}{0.92}
\begin{tabularx}{\textwidth}{@{}p{0.23\textwidth}*{8}{>{\centering\arraybackslash}X}@{}}
\toprule
\multirow{2}{*}{Method} & \multirow{2}{*}{Emb.\,PT.} & \multicolumn{5}{c}{LIBERO} & \multirow{2}{*}{CALVIN} & \multirow{2}{*}{RoboCasa} \\
\cmidrule(lr){3-7}
& & Spatial & Object & Goal & Long & Avg. & & \\
\midrule
\multicolumn{9}{c}{\textit{VLA-based methods}} \\
\midrule
OpenVLA~\citep{kim2024openvlaopensourcevisionlanguageactionmodel} & $\checkmark$ & 84.70 & 88.40 & 79.20 & 53.70 & 76.50 & 3.27 & -- \\
OpenVLA-OFT~\citep{kim2025finetuningvisionlanguageactionmodelsoptimizing} & $\checkmark$ & 97.60 & \underline{98.40} & \underline{97.90} & \underline{94.50} & \underline{97.10} & 4.28 & 33.1 \\
$\pi_{0.5}$~\citep{intelligence2025pi05visionlanguageactionmodelopenworld} & $\checkmark$ & \textbf{98.80} & 98.20 & \textbf{98.00} & 92.40 & 96.90 & 3.97 & \textbf{52.9} \\
CoT-VLA~\citep{zhao2025cotvlavisualchainofthoughtreasoning} & $\checkmark$ & 87.50 & 91.60 & 87.60 & 69.00 & 83.93 & -- & -- \\
GR00T-N1~\citep{nvidia2025gr00tn1openfoundation} & $\checkmark$ & 94.40 & 97.60 & 93.00 & 90.60 & 93.90 & 4.01 & \underline{49.6} \\
X-VLA~\citep{zheng2025xvlasoftpromptedtransformerscalable} & $\checkmark$ & \underline{98.20} & \textbf{98.60} & 97.80 & \textbf{97.60} & \textbf{98.10} & \underline{4.43} & -- \\
FLOWER~\citep{reuss2025flowerdemocratizinggeneralistrobot} & $\checkmark$ & 97.10 & 96.70 & 95.60 & 93.50 & 95.70 & \textbf{4.53} & -- \\
\midrule
\multicolumn{9}{c}{\textit{WAM-based methods}} \\
\midrule
DreamVLA~\citep{zhang2025dreamvlavisionlanguageactionmodeldreamed} & $\checkmark$ & 97.50 & 94.00 & 89.50 & 89.50 & 92.60 & \underline{4.44} & -- \\
UVA~\citep{li2025unifiedvideoactionmodel} & $\times$ & -- & -- & -- & 90.00 & -- & -- & \underline{50.0} \\
LingBot-VA~\citep{li2026causalworldmodelingrobot} & $\checkmark$ & 98.50 & \underline{99.60} & 97.20 & \underline{98.50} & 98.50 & -- & -- \\
Faster-WAM~\citep{ma2026fasterwamworldactionmodels} & $\times$ & \underline{99.10} & 99.50 & 98.40 & 96.20 & 98.30 & -- & -- \\
ST-WAM~\citep{wang2026stwamsemantictemporalworldaction} & $\times$ & 99.00 & \textbf{100.00} & \underline{99.00} & 96.80 & \underline{98.70} & -- & -- \\
SG-WAM~\citep{he2026sgwamtextgroundedspatialawaresemantic} & $\times$ & \textbf{99.40} & 99.40 & 98.20 & 97.80 & \underline{98.70} & -- & -- \\
GaussianWAM~\citep{zhang2026gaussianwamdistillinggeometrysemantics} & $\times$ & -- & -- & -- & -- & 97.60 & -- & -- \\
Fast-WAM~\citep{yuan2026fastwamworldactionmodels} & $\times$ & 98.20 & \textbf{100.00} & 97.00 & 95.20 & 97.60 & 4.13 & 42.3 \\
\midrule
\rowcolor{resultblue}
\textbf{\method{}} & $\times$ & 98.60 & 99.40 & \textbf{99.40} & \textbf{98.60} & \textbf{99.00} & \textbf{4.54} & \textbf{61.4} \\
\multicolumn{1}{@{}l}{$\Delta$ vs.\ baseline} & & +0.40 & -0.60 & +2.40 & +3.40 & +1.40 & +0.41 & +19.1 \\
\bottomrule
\end{tabularx}
\end{table*}

\begin{table}[!t]
\centering
\caption{\textbf{Zero-shot generalization.} LIBERO-Pro averages four shifts. Ext. prior: auxiliary semantic/geometric teachers.}
\label{tab:generalization}
\footnotesize
\setlength{\tabcolsep}{2.5pt}
\setlength{\aboverulesep}{1pt}
\setlength{\belowrulesep}{1pt}
\renewcommand{\arraystretch}{0.95}
\begin{tabularx}{\linewidth}{@{}>{\raggedright\arraybackslash}Xcccc@{}}
\toprule
Method & \shortstack{Emb.\\PT.} & \shortstack{Ext.\\prior} & \shortstack{LIBERO-\\Plus} & \shortstack{LIBERO-\\Pro} \\
\midrule
\multicolumn{5}{c}{\textit{VLA-based methods}} \\
\midrule
OpenVLA~\citep{kim2024openvlaopensourcevisionlanguageactionmodel} & $\checkmark$ & $\times$ & 15.60 & \underline{47.56} \\
OpenVLA-OFT~\citep{kim2025finetuningvisionlanguageactionmodelsoptimizing} & $\checkmark$ & $\times$ & 69.60 & 23.25 \\
$\pi_0$~\citep{black2026pi0visionlanguageactionflowmodel} & $\checkmark$ & $\times$ & 53.60 & 45.25 \\
$\pi_{0.5}$~\citep{intelligence2025pi05visionlanguageactionmodelopenworld} & $\checkmark$ & $\times$ & \textbf{85.70} & \textbf{53.31} \\
X-VLA~\citep{zheng2025xvlasoftpromptedtransformerscalable} & $\checkmark$ & $\times$ & \underline{71.40} & 45.81 \\
NORA~\citep{hung2025norasmallopensourcedgeneralist} & $\checkmark$ & $\times$ & 39.00 & 39.19 \\
\midrule
\multicolumn{5}{c}{\textit{WAM-based methods}} \\
\midrule
LingBot-VA~\citep{li2026causalworldmodelingrobot} & $\checkmark$ & $\times$ & 69.50 & -- \\
Faster-WAM~\citep{ma2026fasterwamworldactionmodels} & $\times$ & $\times$ & 78.30 & -- \\
BridgeWAM~\citep{fang2026bridgewambridgingvideoaction} & $\times$ & $\times$ & 72.24 & \underline{46.48} \\
ST-WAM~\citep{wang2026stwamsemantictemporalworldaction} & $\times$ & $\checkmark$ & 72.80 & -- \\
SG-WAM~\citep{he2026sgwamtextgroundedspatialawaresemantic} & $\times$ & $\checkmark$ & \textbf{81.30} & -- \\
GaussianWAM~\citep{zhang2026gaussianwamdistillinggeometrysemantics} & $\times$ & $\checkmark$ & 71.29 & -- \\
Fast-WAM~\citep{yuan2026fastwamworldactionmodels} & $\times$ & $\times$ & 51.50 & 42.38 \\
\midrule
\rowcolor{resultblue}
\textbf{\method{}} & $\times$ & $\times$ & \underline{78.60} & \textbf{48.99} \\
\multicolumn{1}{@{}l}{$\Delta$ vs.\ baseline} & & & +27.10 & +6.61 \\
\bottomrule
\end{tabularx}
\end{table}

\begin{table}[!t]
\centering
\setlength{\abovecaptionskip}{2pt}
\caption{\textbf{Long-horizon planning.} $\dagger$ indicates the use of privileged simulator state.}
\label{tab:long-horizon}
\footnotesize
\setlength{\tabcolsep}{3pt}
\setlength{\aboverulesep}{1pt}
\setlength{\belowrulesep}{1pt}
\renewcommand{\arraystretch}{0.98}
\begin{tabularx}{\linewidth}{@{}>{\raggedright\arraybackslash}p{0.40\linewidth}*{3}{>{\centering\arraybackslash}X}@{}}
\toprule
\multirow{2}{*}{Method} & Atomic & \multicolumn{2}{c@{}}{Composite} \\
\cmidrule(lr){2-2}\cmidrule(lr){3-4}
& Seen & Seen & Unseen \\
\midrule
Fast-WAM & \underline{16.48} & 0.63 & 2.50 \\
Fast-WAM + VLM planner & -- & 0.83 & 0.21 \\
Fast-WAM + Oracle planner$^{\dagger}$ & -- & \underline{5.63} & \underline{2.71} \\
\midrule
\rowcolor{resultblue}
\textbf{\method{}} & \textbf{29.63} & \textbf{20.42} & \textbf{25.42} \\
\multicolumn{1}{@{}l}{$\Delta$ vs.\ baseline} & +13.15 & +19.79 & +22.92 \\
\bottomrule
\end{tabularx}
\end{table}

\subsection{Experimental Setup}
\label{sec:exp-setup}
\paragraph{Benchmarks.}
We organize evaluation around three capabilities as follows. For \emph{control performance}, we train separate policies on LIBERO~\citep{liu2023liberobenchmarkingknowledgetransfer}, CALVIN~\citep{mees2022calvinbenchmarklanguageconditionedpolicy}, and RoboCasa-Kitchen~\citep{nasiriany2024robocasalargescalesimulationeveryday}. LIBERO comprises 40 tasks across Spatial, Object, Goal, and Long, with 50 trials per task. CALVIN follows ABC$\rightarrow$D, evaluating 1,000 five-instruction chains in unseen environment D and reporting average successful chain length (ACL). RoboCasa-Kitchen comprises 24 tasks, with 50 evaluation trials per task. For \emph{zero-shot generalization}, we apply the LIBERO policy without fine-tuning to LIBERO-Plus~\citep{fei2025liberoplusindepthrobustnessanalysis} and LIBERO-Pro~\citep{zhou2026liberoprorobustfairevaluation}, respectively. Plus covers seven visual and language perturbations; Pro covers Object, Position, Semantic, and Task variations. For \emph{long-horizon planning}, we use RoboCasa365~\citep{nasiriany2026robocasa365largescalesimulationframework}, whose composite tasks require the robot to execute and transition between multiple subtasks. We use Atomic-Seen (18 tasks) as an atomic-skill reference and Composite-Seen and Composite-Unseen (16 tasks each) for long-horizon evaluation, with 30 trials per task. Detailed training and evaluation settings are provided in Appendix~\ref{sec:appendix}.

\paragraph{Implementation.}
We initialize the VLM with Qwen3-VL-4B-Instruct~\citep{bai2025qwen3vltechnicalreport} and the video expert with Wan2.2-TI2V-5B~\citep{wan2025wanopenadvancedlargescale}, connected through a latent interface with $8$ intent tokens by default. Unless otherwise specified, training uses AdamW~\citep{loshchilov2019decoupledweightdecayregularization} with a global batch size of 64 and a cosine learning-rate schedule. Except for the RoboCasa365 setting detailed in Appendix~\ref{sec:robocasa365-protocol}, Stage 1 runs for 10k steps at $10^{-5}$ and Stage 2 for 60k steps at $5\times10^{-5}$. At inference time, we first denoise the goal to a noise level of $\sigma=0.5$ using 5 steps, and then fully denoise the actions using 10 steps. Additional settings are provided in Appendix~\ref{sec:appendix}.

\subsection{Main Results}
\label{sec:exp-main}

\paragraph{Control performance.}
\method{} achieves strong control performance across all three benchmarks without additional embodied pretraining (Table~\ref{tab:main}). It reaches \textbf{99.00\%} average success on LIBERO, an ACL of \textbf{4.54} on CALVIN, and \textbf{61.4\%} success on RoboCasa-Kitchen, the highest aggregate scores among the compared methods. Compared with our Fast-WAM baseline, \method{} improves average success on LIBERO by 1.40 percentage points, CALVIN ACL by 0.41, and success on RoboCasa-Kitchen by 19.1 percentage points. The improvements also extend to long-horizon sequential control: LIBERO-Long success increases from 95.20\% to 98.60\%, while completion of five-task CALVIN chains rises from 71.9\% to 81.5\%. On LIBERO-Long, \method{} even achieves a slightly higher success rate than LingBot-VA~\citep{li2026causalworldmodelingrobot} (98.60\% vs.\ 98.50\%), which incorporates observation history and additional embodied pretraining. These results demonstrate effective manipulation and reliable sequential instruction following across diverse environments.

\paragraph{Zero-shot generalization.}
\label{sec:exp-generalization}
\method{} substantially improves zero-shot generalization over the Fast-WAM baseline and remains competitive with methods using additional embodied pretraining or supervision from external visual priors. Without fine-tuning, the same LIBERO-trained policy achieves \textbf{78.60\%} on LIBERO-Plus and \textbf{48.99\%} on LIBERO-Pro, improving over Fast-WAM by 27.10 and 6.61 percentage points, respectively, and surpassing embodied-pretrained X-VLA on both benchmarks. On LIBERO-Plus, it surpasses GaussianWAM (71.29\%) and ST-WAM (72.80\%) and approaches SG-WAM (81.30\%), all of which incorporate external visual priors through semantic or geometric supervision. These results suggest that our approach of jointly training the VLM and WAM effectively integrates VLM semantic priors with visual prediction and action generation, leading to stronger zero-shot generalization.

\paragraph{Long-horizon planning.}
\label{sec:exp-long-horizon}
\method{} enables the WAM to plan and execute long-horizon tasks (Table~\ref{tab:long-horizon}). It achieves success rates of \textbf{20.42\%} and \textbf{25.42\%} on Composite-Seen and Composite-Unseen, compared with 0.63\% and 2.50\% for the Fast-WAM baseline, respectively. Atomic-Seen success also improves from 16.48\% to \textbf{29.63\%}. To account for the additional subtask supervision used during training, we train a separate Fast-WAM skill policy for each composite suite on annotated subtask segments and evaluate each policy with VLM and oracle schedulers. The VLM planner reaches 0.83\% and 0.21\% on the two composite suites, highlighting the advantage of learning planning and execution jointly over connecting separately trained components. We further construct an oracle subtask scheduler using privileged state information from the simulator to select and schedule subtasks for the same Fast-WAM skill policy. Oracle scheduling achieves 5.63\% and 2.71\% on Composite-Seen and Composite-Unseen, respectively, trailing \method{} by 14.79 and 22.71 percentage points. These results show that replacing task text with subtask descriptions for text supervision and jointly training the VLM and WAM can equip the model with long-horizon planning capabilities. Training and evaluation settings are detailed in Appendix~\ref{sec:robocasa365-protocol}.

\subsection{Ablation Studies}
\label{sec:exp-ablation}

To evaluate the contributions of our design choices, we ablate the model architecture, training strategy, and latent interface supervision on LIBERO and LIBERO-Plus (Table~\ref{tab:ablations}). The full model achieves the best performance on both benchmarks, with larger gains over the ablated variants on LIBERO-Plus.

\begin{table}[!t]
\centering
\caption{\textbf{Ablation studies.} Success rates (\%) on LIBERO and LIBERO-Plus.}
\label{tab:ablations}
\small
\setlength{\tabcolsep}{3pt}
\renewcommand{\arraystretch}{1.12}
\begin{tabularx}{\linewidth}{@{}>{\raggedright\arraybackslash}Xcc@{}}
\toprule
Variant & LIBERO & LIBERO-Plus \\
\midrule
\rowcolor{resultblue}
CoT-WAM \textbf{(ours)} & \textbf{99.00} & \textbf{78.60} \\
\midrule
\multicolumn{3}{c}{\textit{Architecture and training schedule}} \\
\midrule
w/o VLM & 98.00 & 68.65 \\
w/o goal branch & 97.80 & 73.65 \\
single-stage training & 98.50 & 76.10 \\
\midrule
\multicolumn{3}{c}{\textit{Supervision on latent interface}} \\
\midrule
w/o text supervision & 98.45 & 76.70 \\
w/o goal supervision & 98.50 & 73.85 \\
\midrule
Fast-WAM (baseline) & 97.60 & 51.50 \\
\bottomrule
\end{tabularx}
\end{table}

\paragraph{Architecture design.}
VLM conditioning and visual goals that express the VLM's intent improve both control performance and generalization. Replacing the VLM with frozen text embeddings reduces success by 9.95 percentage points on LIBERO-Plus, the largest drop among the evaluated ablations, and by 1.00 point on LIBERO. These results suggest that VLM conditioning strengthens generalization across semantic and visual variations while also improving control performance. Removing the entire goal branch reduces success by 1.20 percentage points on LIBERO and 4.95 points on LIBERO-Plus, further supporting the benefit of grounding the VLM's intent in explicit visual states.

\paragraph{Training strategy.}
We further evaluate end-to-end training, in which both the VLM and WAM are trainable and jointly optimized in a single stage. This setting yields slightly lower performance than our staged training strategy. This supports explicitly supervising the information conveyed through the latent interface before training the WAM to generate actions conditioned on the VLM's intent. The modest performance gap also suggests that our architecture is amenable to joint optimization of the two models from the outset.

\paragraph{Latent interface supervision.}
We evaluate the roles of task text and goal image supervision in Stage 1 by removing each loss separately. Removing the text loss reduces success by 0.55 percentage points on LIBERO and 1.90 points on LIBERO-Plus. This suggests that, beyond enabling long-horizon planning through subtask descriptions, text supervision also improves control performance and zero-shot generalization on the shorter tasks evaluated here. Removing the goal image loss reduces success by 0.50 percentage points on LIBERO and 4.75 points on LIBERO-Plus. This decline, despite retaining goal supervision in Stage 2, highlights the value of grounding VLM intent in visual goals during the initial stage of interface learning.

\subsection{Qualitative Analysis}
\label{sec:visual-grounding}
In this section, we qualitatively analyze how jointly training the VLM and WAM supports scene and semantic generalization and subtask planning in long-horizon tasks. We also visualize successive predicted goals alongside the corresponding execution outcomes, showing how the policy progresses from the initial state to successful task completion.

\begin{figure}[t]
\centering
\includegraphics[width=\linewidth]{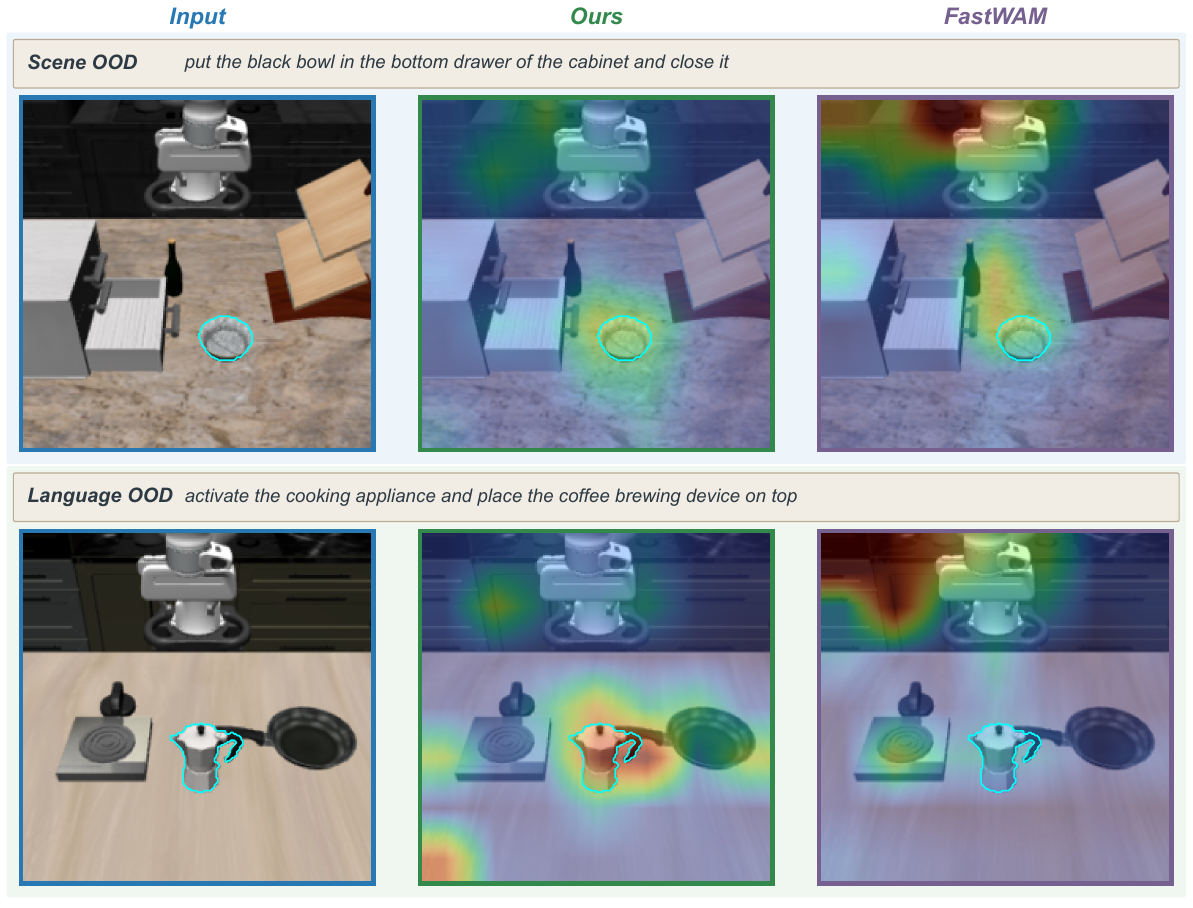}
\caption{\textbf{Attention maps under scene and language shifts.} Comparison on LIBERO-Plus under scene OOD (top) and language OOD (bottom).}
\label{fig:ood-attention}
\end{figure}

\paragraph{Scene and semantic generalization.}
Figure~\ref{fig:ood-attention} compares \method{} and Fast-WAM under scene and linguistic shifts on LIBERO-Plus. In the scene OOD example, \method{} concentrates attention around the requested bowl despite changes in scene appearance. In the language OOD example, it attends to the moka pot described as a ``coffee brewing device.'' By contrast, Fast-WAM exhibits stronger attention responses in background regions in both cases. These examples illustrate task-relevant grounding under scene and language shifts, complementing the quantitative generalization results in Table~\ref{tab:generalization}.

\begin{figure}[!ht]
\centering
\includegraphics[width=0.90\linewidth]{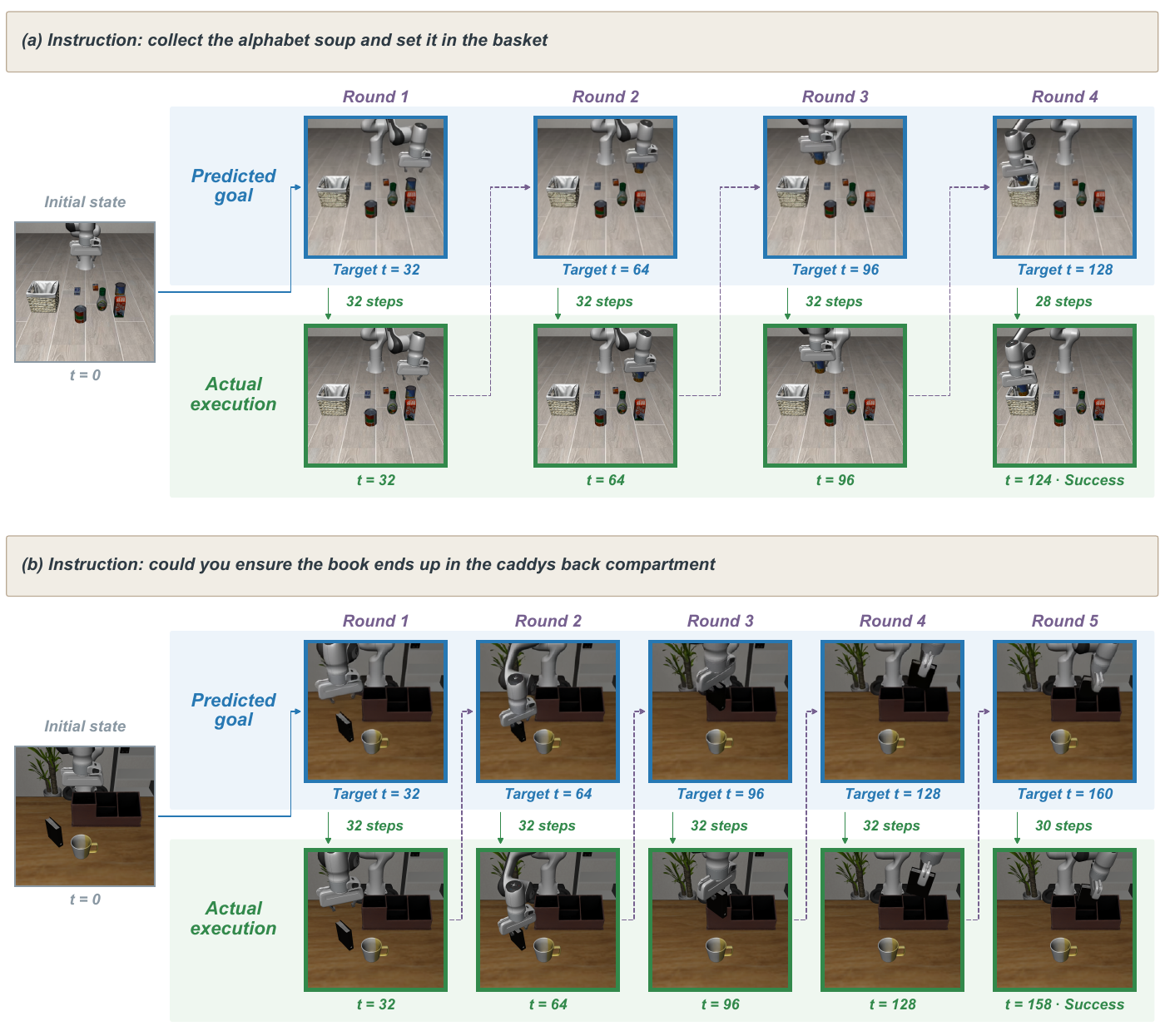}
\caption{\textbf{Visual goals and execution.} Successive predicted visual goals (top) and corresponding execution outcomes (bottom), from the initial state to successful task completion.}
\label{fig:goal-evidence}
\end{figure}

\paragraph{Visual goals and execution.}
Figure~\ref{fig:goal-evidence} compares successive predicted goals with actual execution outcomes over 32-step action chunks, from the initial state to successful task completion. The predicted goals anticipate object lifting and transport, while the corresponding execution outcomes show the robot progressing toward these visual targets. These sequences illustrate how \method{} grounds latent intent in concrete visual states and translates it into action execution.

\begin{figure}[!htbp]
\centering
\includegraphics[width=0.90\linewidth]{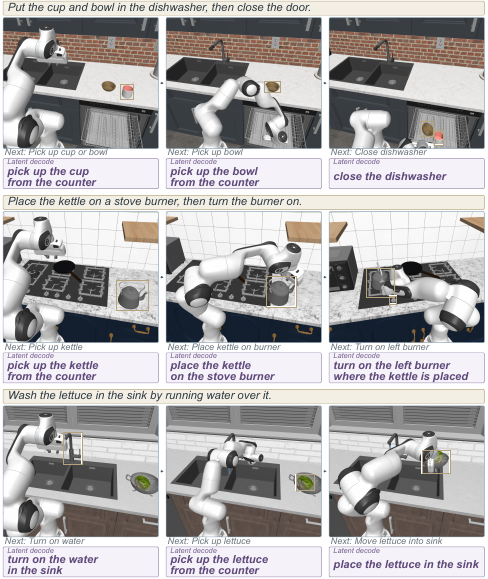}
\caption{\textbf{Subtask planning in long-horizon tasks.}
Subtask predictions across three RoboCasa365 tasks, with the overall
instruction fixed within each task. ``Next'' denotes annotated valid
subtasks; ``Latent decode'' shows predictions decoded from the VLM's
latent queries.}
\label{fig:subtask-prediction}
\end{figure}

\paragraph{Subtask planning.}
Figure~\ref{fig:subtask-prediction} visualizes decoded subtask predictions across three RoboCasa365 tasks on held-out trajectories. With the overall instruction fixed within each task, the predictions adapt to scene changes and align with valid next subtasks. These examples illustrate how \method{} encodes subtask planning intent in its latent interface based on the current scene and task objective.

\subsection{Real-World Experiments}
\label{sec:exp-real-world}

\paragraph{Experimental settings.}
We evaluate \method{} on a bimanual platform equipped with two seven-DoF Marvin M6 robotic arms. The policy uses head and wrist camera images to predict end-effector pose targets and gripper commands. We train a separate policy for each task using 100 demonstrations collected on the robot. Task descriptions and additional implementation details are provided in Appendix~\ref{sec:appendix-real-world}.

\paragraph{Results.}
Table~\ref{tab:real-world} reports success rates over 40 trials per task. \method{} achieves an average success rate of 61.3\%, compared with 47.5\% for Fast-WAM. These results further demonstrate the effectiveness of \method{} for real-world manipulation tasks.

\begin{table}[!htbp]
\centering
\caption{\textbf{Real-world control performance.}}
\label{tab:real-world}
\small
\setlength{\tabcolsep}{4pt}
\renewcommand{\arraystretch}{1.08}
\begin{tabularx}{\linewidth}{@{}>{\raggedright\arraybackslash}Xcc@{}}
\toprule
Task & Fast-WAM & \textbf{\method{}} \\
\midrule
Produce sorting & 52.5 & \textbf{67.5} \\
Table wiping & 42.5 & \textbf{55.0} \\
\midrule
Average & 47.5 & \textbf{61.3} \\
\bottomrule
\end{tabularx}
\end{table}

\vspace{-7pt}
\section{Conclusion}
\label{sec:conclusion}
In this work, we introduced \method{}, a framework that connects VLM semantic reasoning with world-action modeling through continuous plan latents. We jointly train this interface in two stages, using text supervision to align the latents with subtask semantics and goal-image prediction to ground their intent in future visual states. These predicted goals then condition both video and action generation, linking semantic planning to predictive control. Experiments across six benchmarks demonstrate state-of-the-art control performance and improved zero-shot generalization and long-horizon planning. Real-world experiments further validate its effectiveness in physical manipulation.
\clearpage
{\small
\bibliographystyle{ieeenat_fullname}
\bibliography{references}

@misc{nvidia2025gr00tn1openfoundation,
      title={GR00T N1: An Open Foundation Model for Generalist Humanoid Robots}, 
      author={NVIDIA and Johan Bjorck and Fernando Castañeda and Nikita Cherniadev and Xingye Da and Runyu Ding and Linxi "Jim" Fan and Yu Fang and Dieter Fox and Fengyuan Hu and Spencer Huang and Joel Jang and Zhenyu Jiang and Jan Kautz and Kaushil Kundalia and Lawrence Lao and Zhiqi Li and Zongyu Lin and Kevin Lin and Guilin Liu and Edith Llontop and Loic Magne and Ajay Mandlekar and Avnish Narayan and Soroush Nasiriany and Scott Reed and You Liang Tan and Guanzhi Wang and Zu Wang and Jing Wang and Qi Wang and Jiannan Xiang and Yuqi Xie and Yinzhen Xu and Zhenjia Xu and Seonghyeon Ye and Zhiding Yu and Ao Zhang and Hao Zhang and Yizhou Zhao and Ruijie Zheng and Yuke Zhu},
      year={2025},
      eprint={2503.14734},
      archivePrefix={arXiv},
      primaryClass={cs.RO},
      url={https://arxiv.org/abs/2503.14734}, 
}

@misc{black2026pi0visionlanguageactionflowmodel,
      title={$\pi_0$: A Vision-Language-Action Flow Model for General Robot Control}, 
      author={Kevin Black and Noah Brown and Danny Driess and Adnan Esmail and Michael Equi and Chelsea Finn and Niccolo Fusai and Lachy Groom and Karol Hausman and Brian Ichter and Szymon Jakubczak and Tim Jones and Liyiming Ke and Sergey Levine and Adrian Li-Bell and Mohith Mothukuri and Suraj Nair and Karl Pertsch and Lucy Xiaoyang Shi and James Tanner and Quan Vuong and Anna Walling and Haohuan Wang and Ury Zhilinsky},
      year={2026},
      eprint={2410.24164},
      archivePrefix={arXiv},
      primaryClass={cs.LG},
      url={https://arxiv.org/abs/2410.24164}, 
}

@misc{brohan2023rt2visionlanguageactionmodelstransfer,
      title={RT-2: Vision-Language-Action Models Transfer Web Knowledge to Robotic Control}, 
      author={Anthony Brohan and Noah Brown and Justice Carbajal and Yevgen Chebotar and Xi Chen and Krzysztof Choromanski and Tianli Ding and Danny Driess and Avinava Dubey and Chelsea Finn and Pete Florence and Chuyuan Fu and Montse Gonzalez Arenas and Keerthana Gopalakrishnan and Kehang Han and Karol Hausman and Alexander Herzog and Jasmine Hsu and Brian Ichter and Alex Irpan and Nikhil Joshi and Ryan Julian and Dmitry Kalashnikov and Yuheng Kuang and Isabel Leal and Lisa Lee and Tsang-Wei Edward Lee and Sergey Levine and Yao Lu and Henryk Michalewski and Igor Mordatch and Karl Pertsch and Kanishka Rao and Krista Reymann and Michael Ryoo and Grecia Salazar and Pannag Sanketi and Pierre Sermanet and Jaspiar Singh and Anikait Singh and Radu Soricut and Huong Tran and Vincent Vanhoucke and Quan Vuong and Ayzaan Wahid and Stefan Welker and Paul Wohlhart and Jialin Wu and Fei Xia and Ted Xiao and Peng Xu and Sichun Xu and Tianhe Yu and Brianna Zitkovich},
      year={2023},
      eprint={2307.15818},
      archivePrefix={arXiv},
      primaryClass={cs.RO},
      url={https://arxiv.org/abs/2307.15818}, 
}

@misc{fang2026bridgewambridgingvideoaction,
      title={BridgeWAM: Bridging Video and Action Experts via Latent Bridge Queries}, 
      author={Jiaxin Fang and Ruitong Liu and Zeyu Lai and Xingyu Wang and Mingyuan Jia and Pingyue Sheng and Weikang Wan and Mengdi Xu and Yuxiang Gao},
      year={2026},
      eprint={2610.14746},
      archivePrefix={arXiv},
      primaryClass={cs.RO},
      url={https://arxiv.org/abs/2610.14746}, 
}

@misc{fei2025liberoplusindepthrobustnessanalysis,
      title={LIBERO-Plus: In-depth Robustness Analysis of Vision-Language-Action Models}, 
      author={Senyu Fei and Siyin Wang and Junhao Shi and Zihao Dai and Jikun Cai and Pengfang Qian and Li Ji and Xinzhe He and Shiduo Zhang and Zhaoye Fei and Jinlan Fu and Jingjing Gong and Xipeng Qiu},
      year={2025},
      eprint={2510.13626},
      archivePrefix={arXiv},
      primaryClass={cs.RO},
      url={https://arxiv.org/abs/2510.13626}, 
}

@misc{hung2025norasmallopensourcedgeneralist,
      title={NORA: A Small Open-Sourced Generalist Vision Language Action Model for Embodied Tasks}, 
      author={Chia-Yu Hung and Qi Sun and Pengfei Hong and Amir Zadeh and Chuan Li and U-Xuan Tan and Navonil Majumder and Soujanya Poria},
      year={2025},
      eprint={2504.19854},
      archivePrefix={arXiv},
      primaryClass={cs.RO},
      url={https://arxiv.org/abs/2504.19854}, 
}

@misc{kim2024openvlaopensourcevisionlanguageactionmodel,
      title={OpenVLA: An Open-Source Vision-Language-Action Model}, 
      author={Moo Jin Kim and Karl Pertsch and Siddharth Karamcheti and Ted Xiao and Ashwin Balakrishna and Suraj Nair and Rafael Rafailov and Ethan Foster and Grace Lam and Pannag Sanketi and Quan Vuong and Thomas Kollar and Benjamin Burchfiel and Russ Tedrake and Dorsa Sadigh and Sergey Levine and Percy Liang and Chelsea Finn},
      year={2024},
      eprint={2406.09246},
      archivePrefix={arXiv},
      primaryClass={cs.RO},
      url={https://arxiv.org/abs/2406.09246}, 
}

@misc{kim2025finetuningvisionlanguageactionmodelsoptimizing,
      title={Fine-Tuning Vision-Language-Action Models: Optimizing Speed and Success}, 
      author={Moo Jin Kim and Chelsea Finn and Percy Liang},
      year={2025},
      eprint={2502.19645},
      archivePrefix={arXiv},
      primaryClass={cs.RO},
      url={https://arxiv.org/abs/2502.19645}, 
}

@misc{li2025unifiedvideoactionmodel,
      title={Unified Video Action Model}, 
      author={Shuang Li and Yihuai Gao and Dorsa Sadigh and Shuran Song},
      year={2025},
      eprint={2503.00200},
      archivePrefix={arXiv},
      primaryClass={cs.RO},
      url={https://arxiv.org/abs/2503.00200}, 
}

@misc{li2026causalworldmodelingrobot,
      title={Causal World Modeling for Robot Control}, 
      author={Lin Li and Qihang Zhang and Yiming Luo and Shuai Yang and Ruilin Wang and Fei Han and Mingrui Yu and Zelin Gao and Nan Xue and Xing Zhu and Yujun Shen and Yinghao Xu},
      year={2026},
      eprint={2601.21998},
      archivePrefix={arXiv},
      primaryClass={cs.CV},
      url={https://arxiv.org/abs/2601.21998}, 
}

@misc{liu2023liberobenchmarkingknowledgetransfer,
      title={LIBERO: Benchmarking Knowledge Transfer for Lifelong Robot Learning}, 
      author={Bo Liu and Yifeng Zhu and Chongkai Gao and Yihao Feng and Qiang Liu and Yuke Zhu and Peter Stone},
      year={2023},
      eprint={2306.03310},
      archivePrefix={arXiv},
      primaryClass={cs.AI},
      url={https://arxiv.org/abs/2306.03310}, 
}

@misc{ma2026fasterwamworldactionmodels,
      title={Faster-WAM: Do World Action Models Need Deep Action Modules?}, 
      author={Liheng Ma and Rui Heng Yang and Amin Abyaneh and George Z. Xue and Behnam Rahmati and Mateo Clemente and Ziwen Hu and Anlin Chen and Tongtong Cao and Zhanguang Zhang and Yingxue Zhang},
      year={2026},
      eprint={2608.02365},
      archivePrefix={arXiv},
      primaryClass={cs.AI},
      url={https://arxiv.org/abs/2608.02365}, 
}

@misc{mees2022calvinbenchmarklanguageconditionedpolicy,
      title={CALVIN: A Benchmark for Language-Conditioned Policy Learning for Long-Horizon Robot Manipulation Tasks}, 
      author={Oier Mees and Lukas Hermann and Erick Rosete-Beas and Wolfram Burgard},
      year={2022},
      eprint={2112.03227},
      archivePrefix={arXiv},
      primaryClass={cs.RO},
      url={https://arxiv.org/abs/2112.03227}, 
}

@misc{nasiriany2024robocasalargescalesimulationeveryday,
      title={RoboCasa: Large-Scale Simulation of Everyday Tasks for Generalist Robots}, 
      author={Soroush Nasiriany and Abhiram Maddukuri and Lance Zhang and Adeet Parikh and Aaron Lo and Abhishek Joshi and Ajay Mandlekar and Yuke Zhu},
      year={2024},
      eprint={2406.02523},
      archivePrefix={arXiv},
      primaryClass={cs.RO},
      url={https://arxiv.org/abs/2406.02523}, 
}

@misc{intelligence2025pi05visionlanguageactionmodelopenworld,
      title={$\pi_{0.5}$: a Vision-Language-Action Model with Open-World Generalization}, 
      author={Physical Intelligence and Kevin Black and Noah Brown and James Darpinian and Karan Dhabalia and Danny Driess and Adnan Esmail and Michael Equi and Chelsea Finn and Niccolo Fusai and Manuel Y. Galliker and Dibya Ghosh and Lachy Groom and Karol Hausman and Brian Ichter and Szymon Jakubczak and Tim Jones and Liyiming Ke and Devin LeBlanc and Sergey Levine and Adrian Li-Bell and Mohith Mothukuri and Suraj Nair and Karl Pertsch and Allen Z. Ren and Lucy Xiaoyang Shi and Laura Smith and Jost Tobias Springenberg and Kyle Stachowicz and James Tanner and Quan Vuong and Homer Walke and Anna Walling and Haohuan Wang and Lili Yu and Ury Zhilinsky},
      year={2025},
      eprint={2504.16054},
      archivePrefix={arXiv},
      primaryClass={cs.LG},
      url={https://arxiv.org/abs/2504.16054}, 
}

@misc{reuss2025flowerdemocratizinggeneralistrobot,
      title={FLOWER: Democratizing Generalist Robot Policies with Efficient Vision-Language-Action Flow Policies}, 
      author={Moritz Reuss and Hongyi Zhou and Marcel Rühle and Ömer Erdinç Yağmurlu and Fabian Otto and Rudolf Lioutikov},
      year={2025},
      eprint={2509.04996},
      archivePrefix={arXiv},
      primaryClass={cs.RO},
      url={https://arxiv.org/abs/2509.04996}, 
}

@misc{yuan2026fastwamworldactionmodels,
      title={Fast-WAM: Do World Action Models Need Test-time Future Imagination?}, 
      author={Tianyuan Yuan and Zibin Dong and Yicheng Liu and Hang Zhao},
      year={2026},
      eprint={2603.16666},
      archivePrefix={arXiv},
      primaryClass={cs.CV},
      url={https://arxiv.org/abs/2603.16666}, 
}

@misc{zhang2025dreamvlavisionlanguageactionmodeldreamed,
      title={DreamVLA: A Vision-Language-Action Model Dreamed with Comprehensive World Knowledge}, 
      author={Wenyao Zhang and Hongsi Liu and Zekun Qi and Yunnan Wang and Xinqiang Yu and Jiazhao Zhang and Runpei Dong and Jiawei He and Fan Lu and He Wang and Zhizheng Zhang and Li Yi and Wenjun Zeng and Xin Jin},
      year={2025},
      eprint={2507.04447},
      archivePrefix={arXiv},
      primaryClass={cs.CV},
      url={https://arxiv.org/abs/2507.04447}, 
}

@misc{zhao2025cotvlavisualchainofthoughtreasoning,
      title={CoT-VLA: Visual Chain-of-Thought Reasoning for Vision-Language-Action Models}, 
      author={Qingqing Zhao and Yao Lu and Moo Jin Kim and Zipeng Fu and Zhuoyang Zhang and Yecheng Wu and Zhaoshuo Li and Qianli Ma and Song Han and Chelsea Finn and Ankur Handa and Ming-Yu Liu and Donglai Xiang and Gordon Wetzstein and Tsung-Yi Lin},
      year={2025},
      eprint={2503.22020},
      archivePrefix={arXiv},
      primaryClass={cs.CV},
      url={https://arxiv.org/abs/2503.22020}, 
}

@misc{zheng2025xvlasoftpromptedtransformerscalable,
      title={X-VLA: Soft-Prompted Transformer as Scalable Cross-Embodiment Vision-Language-Action Model}, 
      author={Jinliang Zheng and Jianxiong Li and Zhihao Wang and Dongxiu Liu and Xirui Kang and Yuchun Feng and Yinan Zheng and Jiayin Zou and Yilun Chen and Jia Zeng and Ya-Qin Zhang and Jiangmiao Pang and Jingjing Liu and Tai Wang and Xianyuan Zhan},
      year={2025},
      eprint={2510.10274},
      archivePrefix={arXiv},
      primaryClass={cs.RO},
      url={https://arxiv.org/abs/2510.10274}, 
}

@misc{zhou2026liberoprorobustfairevaluation,
      title={LIBERO-PRO: Towards Robust and Fair Evaluation of Vision-Language-Action Models Beyond Memorization}, 
      author={Xueyang Zhou and Yangming Xu and Guiyao Tie and Yongchao Chen and Guowen Zhang and Duanfeng Chu and Pan Zhou and Lichao Sun},
      year={2026},
      eprint={2510.03827},
      archivePrefix={arXiv},
      primaryClass={cs.CV},
      url={https://arxiv.org/abs/2510.03827}, 
}

@misc{zhu2026dswamdualsystemworldaction,
      title={DSWAM: A Dual-System World Action Foundation Model for Fine-Grained Robot Manipulation}, 
      author={Jian Zhu and Jianjun Zhang and Taiyi Su and Tianbin Liu and Zhangyuan Wang and Kai Xie and Zitai Huang and Chong Ma and Youzhang He and Tianjian Wang and Hanyang Wang and Weihao Ding and Yi Xu},
      year={2026},
      eprint={2607.04927},
      archivePrefix={arXiv},
      primaryClass={cs.RO},
      url={https://arxiv.org/abs/2607.04927}, 
}

@misc{loshchilov2019decoupledweightdecayregularization,
      title={Decoupled Weight Decay Regularization}, 
      author={Ilya Loshchilov and Frank Hutter},
      year={2019},
      eprint={1711.05101},
      archivePrefix={arXiv},
      primaryClass={cs.LG},
      url={https://arxiv.org/abs/1711.05101}, 
}

@misc{lipman2023flowmatchinggenerativemodeling,
      title={Flow Matching for Generative Modeling}, 
      author={Yaron Lipman and Ricky T. Q. Chen and Heli Ben-Hamu and Maximilian Nickel and Matt Le},
      year={2023},
      eprint={2210.02747},
      archivePrefix={arXiv},
      primaryClass={cs.LG},
      url={https://arxiv.org/abs/2210.02747}, 
}

@misc{ye2026worldactionmodelszeroshot,
      title={World Action Models are Zero-shot Policies}, 
      author={Seonghyeon Ye and Yunhao Ge and Kaiyuan Zheng and Shenyuan Gao and Sihyun Yu and George Kurian and Suneel Indupuru and You Liang Tan and Chuning Zhu and Jiannan Xiang and Ayaan Malik and Kyungmin Lee and William Liang and Nadun Ranawaka and Jiasheng Gu and Yinzhen Xu and Guanzhi Wang and Fengyuan Hu and Avnish Narayan and Johan Bjorck and Jing Wang and Gwanghyun Kim and Dantong Niu and Ruijie Zheng and Yuqi Xie and Jimmy Wu and Qi Wang and Ryan Julian and Danfei Xu and Yilun Du and Yevgen Chebotar and Scott Reed and Jan Kautz and Yuke Zhu and Linxi "Jim" Fan and Joel Jang},
      year={2026},
      eprint={2602.15922},
      archivePrefix={arXiv},
      primaryClass={cs.RO},
      url={https://arxiv.org/abs/2602.15922}, 
}

@misc{wang2026stwamsemantictemporalworldaction,
      title={ST-WAM: Semantic-Temporal World Action Model for Robust Manipulation under Visual Distribution Shifts}, 
      author={Mingxin Wang and Bin Hu and Bin Qian and Kaitao Jiang and Haoning Wu and Feng Yan and Bowen Jing and Ruiyang Hao and Enyi Wang and Kangning Niu and Yandan Yang and Mu Xu and Yan Wang and Houde Liu and Tianlun Li},
      year={2026},
      eprint={2607.28993},
      archivePrefix={arXiv},
      primaryClass={cs.RO},
      url={https://arxiv.org/abs/2607.28993}, 
}

@misc{he2026sgwamtextgroundedspatialawaresemantic,
      title={SG-WAM: Text-Grounded and Spatial-aware Semantic Guidance for World-Action Models}, 
      author={Junjie He and Junfeng Li and Zhide Zhong and Haodong Yan and Ruixin Li and Yangyang Zheng and Jiaguan Zhu and Tianran Zhang and Yuqiao Du and Wen Chen and Shunbo Zhou and Haoang Li},
      year={2026},
      eprint={2608.08839},
      archivePrefix={arXiv},
      primaryClass={cs.RO},
      url={https://arxiv.org/abs/2608.08839}, 
}

@misc{zhang2026gaussianwamdistillinggeometrysemantics,
      title={GaussianWAM: Distilling Geometry and Semantics from 3D Gaussian Fields into World-Action Models}, 
      author={Zijian Zhang and Yuqing Jiang and Weitao Zhou and Minglei Li and Jinhao Zhang and Yao Mu and Xiaofan Li and Hao Zhao and Haibao Yu},
      year={2026},
      eprint={2608.24714},
      archivePrefix={arXiv},
      primaryClass={cs.RO},
      url={https://arxiv.org/abs/2608.24714}, 
}

@misc{bai2025qwen3vltechnicalreport,
      title={Qwen3-VL Technical Report}, 
      author={Shuai Bai and Yuxuan Cai and Ruizhe Chen and Keqin Chen and Xionghui Chen and Zesen Cheng and Lianghao Deng and Wei Ding and Chang Gao and Chunjiang Ge and Wenbin Ge and Zhifang Guo and Qidong Huang and Jie Huang and Fei Huang and Binyuan Hui and Shutong Jiang and Zhaohai Li and Mingsheng Li and Mei Li and Kaixin Li and Zicheng Lin and Junyang Lin and Xuejing Liu and Jiawei Liu and Chenglong Liu and Yang Liu and Dayiheng Liu and Shixuan Liu and Dunjie Lu and Ruilin Luo and Chenxu Lv and Rui Men and Lingchen Meng and Xuancheng Ren and Xingzhang Ren and Sibo Song and Yuchong Sun and Jun Tang and Jianhong Tu and Jianqiang Wan and Peng Wang and Pengfei Wang and Qiuyue Wang and Yuxuan Wang and Tianbao Xie and Yiheng Xu and Haiyang Xu and Jin Xu and Zhibo Yang and Mingkun Yang and Jianxin Yang and An Yang and Bowen Yu and Fei Zhang and Hang Zhang and Xi Zhang and Bo Zheng and Humen Zhong and Jingren Zhou and Fan Zhou and Jing Zhou and Yuanzhi Zhu and Ke Zhu},
      year={2025},
      eprint={2511.21631},
      archivePrefix={arXiv},
      primaryClass={cs.CV},
      url={https://arxiv.org/abs/2511.21631}, 
}

@misc{li2026wallwmcarvingworldaction,
      title={WALL-WM: Carving World Action Modeling at the Event Joints}, 
      author={Shalfun Li and Victor Yao and Charles Yang and Truth Qu and Regis Cheng and Ryan Yu and Howard Lu and Newton Von and Vincent Chen and Yohann Tang and Maeve Zhang and Ellie Ma and Gody Li and Starrick Liu and Sage Yang and Lorien Shu and J. W. Gao and Ethan Chen and Colin Ye and Yu Sun and Elise Mon and PS Zhang and Neo Li and Lily Li and James Wang and Ping Yang and Chris Pan and Lucy Liang and Hang Su and Roy Gan and Hao Wang and Qian Wang},
      year={2026},
      eprint={2606.01955},
      archivePrefix={arXiv},
      primaryClass={cs.RO},
      url={https://arxiv.org/abs/2606.01955}, 
}

@misc{wan2025wanopenadvancedlargescale,
      title={Wan: Open and Advanced Large-Scale Video Generative Models}, 
      author={Team Wan and Ang Wang and Baole Ai and Bin Wen and Chaojie Mao and Chen-Wei Xie and Di Chen and Feiwu Yu and Haiming Zhao and Jianxiao Yang and Jianyuan Zeng and Jiayu Wang and Jingfeng Zhang and Jingren Zhou and Jinkai Wang and Jixuan Chen and Kai Zhu and Kang Zhao and Keyu Yan and Lianghua Huang and Mengyang Feng and Ningyi Zhang and Pandeng Li and Pingyu Wu and Ruihang Chu and Ruili Feng and Shiwei Zhang and Siyang Sun and Tao Fang and Tianxing Wang and Tianyi Gui and Tingyu Weng and Tong Shen and Wei Lin and Wei Wang and Wei Wang and Wenmeng Zhou and Wente Wang and Wenting Shen and Wenyuan Yu and Xianzhong Shi and Xiaoming Huang and Xin Xu and Yan Kou and Yangyu Lv and Yifei Li and Yijing Liu and Yiming Wang and Yingya Zhang and Yitong Huang and Yong Li and You Wu and Yu Liu and Yulin Pan and Yun Zheng and Yuntao Hong and Yupeng Shi and Yutong Feng and Zeyinzi Jiang and Zhen Han and Zhi-Fan Wu and Ziyu Liu},
      year={2025},
      eprint={2503.20314},
      archivePrefix={arXiv},
      primaryClass={cs.CV},
      url={https://arxiv.org/abs/2503.20314}, 
}

@misc{nasiriany2026robocasa365largescalesimulationframework,
      title={RoboCasa365: A Large-Scale Simulation Framework for Training and Benchmarking Generalist Robots}, 
      author={Soroush Nasiriany and Sepehr Nasiriany and Abhiram Maddukuri and Yuke Zhu},
      year={2026},
      eprint={2603.04356},
      archivePrefix={arXiv},
      primaryClass={cs.RO},
      url={https://arxiv.org/abs/2603.04356}, 
}

@misc{chung2023unimaxfairereffectivelanguage,
      title={UniMax: Fairer and more Effective Language Sampling for Large-Scale Multilingual Pretraining}, 
      author={Hyung Won Chung and Noah Constant and Xavier Garcia and Adam Roberts and Yi Tay and Sharan Narang and Orhan Firat},
      year={2023},
      eprint={2304.09151},
      archivePrefix={arXiv},
      primaryClass={cs.CL},
      url={https://arxiv.org/abs/2304.09151}, 
}

@misc{liang2025mixtureoftransformerssparsescalablearchitecture,
      title={Mixture-of-Transformers: A Sparse and Scalable Architecture for Multi-Modal Foundation Models}, 
      author={Weixin Liang and Lili Yu and Liang Luo and Srinivasan Iyer and Ning Dong and Chunting Zhou and Gargi Ghosh and Mike Lewis and Wen-tau Yih and Luke Zettlemoyer and Xi Victoria Lin},
      year={2025},
      eprint={2411.04996},
      archivePrefix={arXiv},
      primaryClass={cs.CL},
      url={https://arxiv.org/abs/2411.04996}, 
}
}

\clearpage
\appendix

\section{Implementation and Evaluation Details}
\label{sec:appendix}

\subsection{Model and Training Configuration}
\label{sec:appendix-implementation}

We initialize the VLM with Qwen3-VL-4B-Instruct and the video expert with Wan2.2-TI2V-5B. For LIBERO, CALVIN, and RoboCasa-Kitchen, the VLM communicates with the WAM through 8 learnable intent tokens and an MLP projection. The action expert has a hidden dimension of 1,024, and the visual VAE remains frozen. Table~\ref{tab:appendix-training} summarizes the configuration for LIBERO, CALVIN, and RoboCasa-Kitchen; the RoboCasa365 configuration is described in Section~\ref{sec:robocasa365-protocol}.

\begin{table}[!ht]
\centering
\caption{\textbf{Training configuration.} Settings for LIBERO, CALVIN, and RoboCasa-Kitchen.}
\label{tab:appendix-training}
\small
\setlength{\tabcolsep}{3pt}
\renewcommand{\arraystretch}{1.08}
\begin{tabularx}{\linewidth}{@{}>{\raggedright\arraybackslash}Xcc@{}}
\toprule
Setting & Stage 1 & Stage 2 \\
\midrule
Training steps & 10k & 60k \\
Learning rate & $10^{-5}$ & $5\times10^{-5}$ \\
Global batch size & 64 & 64 \\
Optimizer & AdamW & AdamW \\
Weight decay & 0.01 & 0.01 \\
Learning-rate schedule & Cosine & Cosine \\
Precision & BF16 & BF16 \\
\midrule
Text loss weight & 0.05 & 0 \\
Goal loss weight & 1 & 1 \\
Video loss weight & 1 & 1 \\
Action loss weight & 0 & 1 \\
\bottomrule
\end{tabularx}
\end{table}

Stage 1 jointly optimizes the VLM, intent queries, projection, and video expert using text, goal-image, and video supervision. Stage 2 loads these weights and initializes the action expert by linearly interpolating the video expert's weights. It then freezes the VLM and conditioning projection while optimizing the video and action experts with goal, video, and action losses.

\paragraph{Goal conditioning and inference.}
During training, the goal condition is clean with probability 0.5. Otherwise, its noise level is sampled uniformly from $[0,0.5]$ and applied according to Eq.~\ref{eq:goal-condition-noise}. At inference, the VLM computes the intent condition in one forward pass. We then denoise the goal for 5 steps to $\sigma=0.5$, hold this latent fixed, and denoise the action chunk for 10 steps. Text decoding and full future-video generation are omitted.

\subsection{Training Windows and Supervision}
\label{sec:training-data}

For LIBERO, CALVIN, and RoboCasa-Kitchen, each training window spans 32 action steps and contains nine images sampled at a stride of four steps, including the current observation and the endpoint. On LIBERO and CALVIN, external and wrist-camera images are resized to $224\times224$ and concatenated horizontally. Actions retain each dataset's native parameterization and are normalized using training-set statistics.

The text target is the task instruction for these three benchmarks. For RoboCasa365 composite tasks, the text target is the annotated subtask description associated with the current state. For Atomic-Seen, the original task instruction serves as the text target. In all settings, target-text tokens attend to the intent queries and their own causal text prefix, while the intent queries aggregate the observation and overall task instruction.

\subsection{Control Performance}
\label{sec:appendix-control}

We train separate policies on LIBERO, CALVIN, and RoboCasa-Kitchen using their respective demonstration datasets and evaluate them with closed-loop replanning. LIBERO covers the Spatial, Object, Goal, and Long suites, while RoboCasa-Kitchen evaluates manipulation in kitchen environments. CALVIN follows the ABC$\rightarrow$D protocol, training on language-annotated segments from environments A, B, and C and evaluating instruction chains in the unseen environment D. We report task success rates for LIBERO and RoboCasa-Kitchen, and average successful chain length and chain-completion rates for CALVIN.

\subsection{Zero-Shot Generalization}
\label{sec:appendix-generalization}

We evaluate the LIBERO-trained policy on LIBERO-Plus and LIBERO-Pro without fine-tuning. LIBERO-Plus tests robustness to visual and language perturbations, while LIBERO-Pro covers Object, Position, Semantic, and Task variations. We report success rates, averaging across variation categories for LIBERO-Pro.

\paragraph{Comparison fields.}
In Tables~\ref{tab:main} and~\ref{tab:generalization}, additional embodied pretraining refers to pretraining on robot data beyond the benchmark-specific training set. The external-prior field in Table~\ref{tab:generalization} denotes auxiliary pretrained semantic or geometric models used to guide or supervise policy learning.

\subsection{Long-Horizon Planning}
\label{sec:robocasa365-protocol}

\paragraph{Task suites and data.}
In RoboCasa365, composite tasks involve mobile manipulation across multiple linked subtasks, requiring the robot to coordinate their execution and transition between them. We evaluate atomic-skill execution on Atomic-Seen and long-horizon task completion on Composite-Seen and Composite-Unseen. These suites contain 18, 16, and 16 tasks, respectively. Each task uses 30 evaluation trials with seeds 0--29 and the official task horizon, giving 540 Atomic-Seen trials and 480 trials in each composite suite.

Standalone Fast-WAM and \method{} policies are trained separately for Atomic-Seen, Composite-Seen, and Composite-Unseen using 10\% of the demonstrations from the \texttt{target} split (50 demonstrations per task): 900 demonstrations for Atomic-Seen and 800 for each composite suite. For \method{}, Atomic-Seen uses the original task description for text supervision and does not involve subtask prediction. Both composite suites use annotated subtask descriptions as text-supervision targets. At inference, the composite policies infer subtask intent through the latent interface from the overall task instruction and current observation.

\paragraph{Training and control settings.}
The \method{} models for Atomic-Seen, Composite-Seen, and Composite-Unseen use 16 intent queries and a 64-step action horizon. Their three camera views are resized to $256\times256$ and concatenated horizontally, and nine images are sampled at a stride of eight control steps. Stage 1 runs for 30k updates at a learning rate of $10^{-5}$, with unit weights for text, goal, and video losses. Stage 2 starts from this checkpoint, uses a learning rate of $10^{-4}$, and is evaluated at 150k updates. The corresponding Fast-WAM and skill-policy checkpoints are also evaluated at 150k updates.

At inference, \method{} follows the goal and action denoising procedure described above. Both \method{} and Fast-WAM predict 64-step action chunks and execute the first 32 actions before replanning.

\paragraph{VLM-scheduled skill policy.}
To account for the subtask annotations used by \method{}, we train separate Fast-WAM skill policies for Composite-Seen and Composite-Unseen, each using annotated subtask windows from 800 demonstrations covering 16 tasks. Each skill policy receives a subtask description and generates the corresponding robot actions. A Qwen3-VL-4B-Instruct planner selects a subtask every 64 control steps from the current scene and overall instruction, while the skill policy replans every 32 steps using ten action-denoising steps. This comparison provides the subtask information through an explicit planner--policy interface.

\paragraph{Oracle-scheduled skill policy.}
We then replace the VLM planner with an oracle subtask scheduler constructed from privileged simulator state, while retaining the same Fast-WAM skill policy. The scheduler uses object positions, grasp and contact states, fixture states, and task-progress predicates to select the next subtask. It checks whether the required state transition has occurred before advancing and rechecks prerequisites after a failed grasp or placement. Subtasks are updated every 64 control steps, and the skill policy continues to generate all robot actions. This setting assigns subtask selection and scheduling to the privileged-state oracle, allowing the comparison to focus on how effectively the learned policy carries out the resulting sequence.

\subsection{Real-World Experiments}
\label{sec:appendix-real-world}

\begin{figure*}[!t]
\centering
\includegraphics[width=\textwidth]{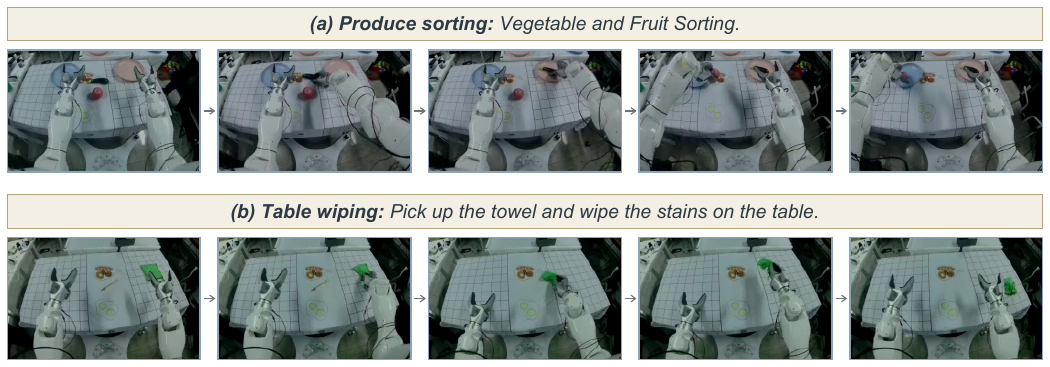}
\caption{\textbf{Real-world task demonstrations.}
Produce sorting (top) and table wiping (bottom).
Each sequence progresses from left to right.}
\label{fig:real-world-execution}
\end{figure*}

Figure~\ref{fig:real-world-execution} illustrates real-world task execution. The platform has two seven-DoF Marvin M6 arms and provides a head-camera view and an upper-wrist view from each arm. We consider the following tasks:
\begin{itemize}
    \setlength{\itemsep}{2pt}
    \setlength{\parskip}{0pt}
    \item \textbf{Produce sorting.} The robot sorts fruits and vegetables onto their designated plates according to their categories.
    \item \textbf{Table wiping.} The robot picks up a towel and wipes the designated stain from the tabletop.
\end{itemize}

\paragraph{Data and action representation.}
For each task, we use 100 demonstrations for training, all collected on our robot platform. The policy predicts 32-step bimanual action chunks. Each arm's action consists of a three-dimensional translation, a six-dimensional rotation representation, and an absolute gripper opening, giving 20 dimensions per bimanual action. For each arm, all predicted end-effector poses are expressed relative to its measured pose at the start of the action chunk.

\section{Ablation Definitions}
\label{sec:appendix-ablations}

The variants in Table~\ref{tab:ablations} examine architecture, training strategy, and supervision on the latent interface. They use the same evaluation episodes and decoding settings as the full model. The full model follows the 10k-step Stage 1 and 60k-step Stage 2 schedule in Table~\ref{tab:appendix-training}.

\paragraph{Architecture.}
The \emph{w/o VLM} variant replaces the observation-conditioned VLM interface with frozen UMT5~\citep{chung2023unimaxfairereffectivelanguage} instruction embeddings supplied directly to the WAM. The \emph{w/o goal branch} variant removes the goal-prediction tokens, goal loss, and goal-conditioning paths. It retains text and video supervision in Stage 1 and video and action supervision in Stage 2.

\paragraph{Training strategy.}
The \emph{single-stage training} variant jointly optimizes the VLM, latent queries, projection, video expert, and action expert for 60k steps. All modules are trainable from the outset, with text, goal, video, and action supervision applied within this single stage.

\paragraph{Supervision on the latent interface.}
The \emph{w/o text supervision} variant removes the text loss in Stage 1 while retaining goal and video supervision. The \emph{w/o goal supervision} variant removes the Stage 1 goal-image loss while retaining the goal branch, text loss, and video loss. Both variants subsequently use the standard Stage 2 objective, including goal, video, and action supervision. Thus, these ablations change interface supervision in Stage 1, whereas removing the goal branch changes the architecture in both stages.

\section{Qualitative Visualization Protocols}
\label{sec:appendix-qualitative}

\subsection{Scene and Semantic Generalization}
\label{sec:attention-protocol}

For Figure~\ref{fig:ood-attention}, both methods receive identical observations and instructions from LIBERO-Plus. We visualize WAM attention from action queries to observation tokens. Each map is normalized to have mean one across both camera views, and the external-camera view is displayed using a shared color range of $[0,4]$. Cyan contours mark the target objects using simulator masks.

\subsection{Visual Goals and Execution}
\label{sec:visual-protocol}

For Figure~\ref{fig:goal-evidence}, we use the 60k-step checkpoint to execute two LIBERO-Plus tasks in 32-step action chunks, replanning from the resulting observation until success. Each column pairs the goal predicted before a chunk with its execution outcome. To display the goal at $\sigma=0.5$, we estimate the clean latent as $\hat{x}_0=x_\sigma-\sigma v_\theta$ and decode it with the VAE, while actions remain conditioned on $x_\sigma$. All panels within a task use the same camera view.

\subsection{Subtask Planning}
\label{sec:subtask-protocol}

For Figure~\ref{fig:subtask-prediction}, we use the RoboCasa365 Composite-Seen 150k-step checkpoint to decode subtasks at three states from each of three held-out trajectories. The overall instruction remains fixed within each task. At each state, the VLM's text head greedily decodes from the intent queries and previously generated tokens. The ``Next'' annotations identify valid next subtasks from simulator state, enabling comparison with the decoded intent.

\section{Limitations}
\label{sec:limit}

The visual goals in \method{} are defined by a fixed prediction horizon, which may not align with semantic subtask boundaries. The framework also lacks an explicit mechanism for assessing whether a predicted goal is achievable from the current state. Future work could explore adaptive goal horizons and goal feasibility assessment to support more flexible planning and recovery from unsuccessful execution.

\end{document}